\documentclass[10pt,twocolumn,letterpaper]{article}

\usepackage{wacv}
\usepackage{times}
\usepackage{epsfig}
\usepackage{graphicx}
\usepackage{amsmath}
\usepackage{amssymb}

\usepackage{multirow}
\usepackage{subfigure}
\usepackage{array}
\usepackage{dblfloatfix}
\usepackage{breqn}
\usepackage{algorithm}
\usepackage{algorithmic}
\usepackage{balance}
\usepackage[table]{xcolor}
\usepackage[breaklinks=true,bookmarks=false]{hyperref}
\newcolumntype{P}[1]{>{\centering\arraybackslash}p{#1}}

\def\wacvPaperID{123} 
\wacvfinalcopy
\begin{document}

\title{Reconstructing Training Data from Diverse ML Models by Ensemble Inversion}

\author{Qian Wang\\
Apple\\
{\tt\small qianwang@apple.com} \and
Daniel Kurz\\
Apple\\
{\tt\small daniel_kurz@apple.com}
}

\maketitle
\thispagestyle{empty}

\begin{abstract}
Model Inversion (MI), in which an adversary abuses access to a trained Machine Learning (ML) model attempting to infer sensitive information about its original training data, has attracted increasing research attention. During MI, the trained model under attack (MUA) is usually frozen and used to guide the training of a generator, such as a Generative Adversarial Network (GAN), to reconstruct the distribution of the original training data of that model. This might cause leakage of original training samples, and if successful, the privacy of dataset subjects will be at risk if the training data contains Personally Identifiable Information (PII). Therefore, an in-depth investigation of the potentials of MI techniques is crucial for the development of corresponding defense techniques. High-quality reconstruction of training data based on a single model is challenging. However, existing MI literature does not explore targeting multiple models jointly, which may provide additional information and diverse perspectives to the adversary.

We propose the ensemble inversion technique that estimates the distribution of original training data by training a generator constrained by an \textbf{ensemble} (or set) of trained models with shared subjects or entities. This technique leads to noticeable improvements of the quality of the generated samples with distinguishable features of the dataset entities compared to MI of a single ML model. We achieve high quality results without any dataset and show how utilizing an auxiliary dataset that's similar to the presumed training data improves the results. The impact of model diversity in the ensemble is thoroughly investigated and additional constraints are utilized to encourage sharp predictions and high activations for the reconstructed samples, leading to more accurate reconstruction of training images.
\end{abstract}

\section{Introduction}
Deep convolutional neural networks (DCNNs) have been successfully used in a wide range of applications related to computer vision, speech recognition, and healthcare. However, significant concerns about privacy are raised by the fact that many of these applications involve processing sensitive and proprietary datasets. In particular, when private or PII data is used for training DCNN models, the obtained models may potentially leak sensitive information about dataset subjects through the model output. 

Model Inversion (MI) attacks aim to extract sensitive features of training data by abusing access to the input and output of a predictive model. Fredrikson \etal~\cite{fredrikson2014privacy} proposed the first MI attack, which infers private genomic attributes about individuals in the training dataset. Fredrikson \etal~\cite{fredrikson2015model} then extended MI attacks to reconstruct faces over logistic regression and decision trees. Kusano and Sakuma~\cite{kusano2018classifier} presented a framework to reconstruct training images by using a Generative Adversarial Network (GAN) and an auxiliary dataset that has a similar distribution as the (presumed) original training data. Recently, a GAN-based method was used to extract PII from a face recognition system~\cite{zhang2020secret}. They demonstrate very realistic face reconstructions because they utilize partially blocked training images with known identities as additional input to their attack.

\begin{figure*}[!t]
\centering
\includegraphics[width=0.8\linewidth]{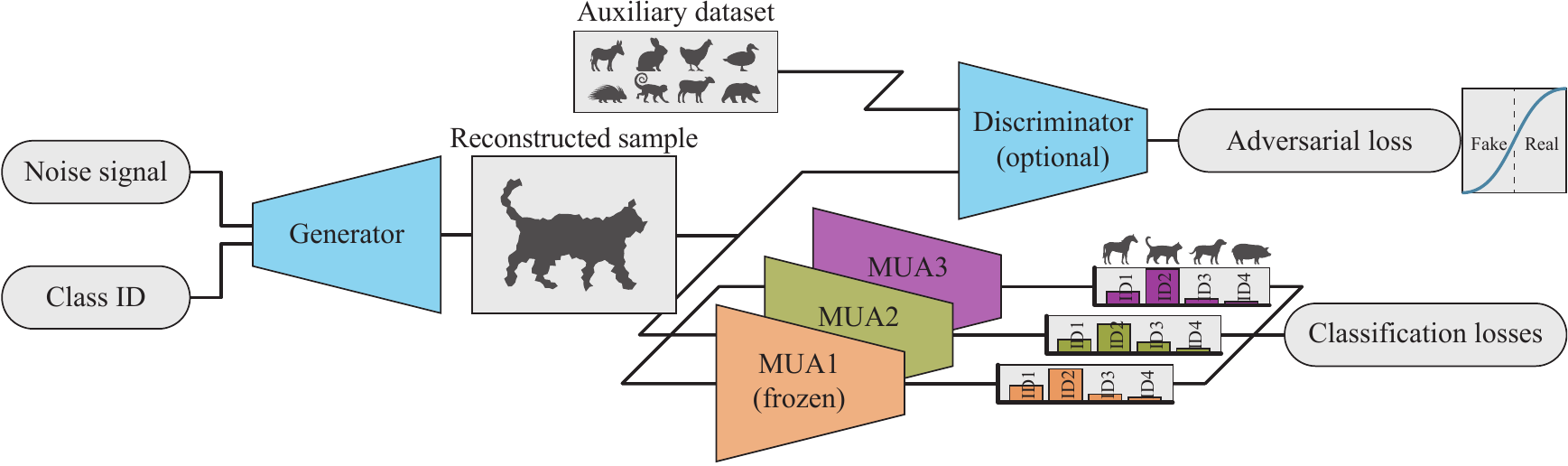}
\caption{Diagram of the proposed method for model inversion based on an \emph{ensemble} of Models Under Attack (MUAs). In order to reconstruct images from the original training data of MUAs, a generator is trained with two types of constraints for extracting useful information from the MUAs. An ensemble of MUAs is used to guide the generator to make all models predict the desired class ID. And a discriminator guides the generator to produce realistic images that look like data from the distribution of an auxiliary dataset (that is assumed to look similar to the original training data), as done when training a regular GAN. If such an auxiliary dataset is not available, we use a data-free setup without the discriminator instead.}
\label{fig:overview}
\end{figure*}

Thus far, in the context of model inversion, recovering the original training images from a trained model with a GAN is attracting more and more research attention. However, how to extract private training data from a model without access to partially blocked training data, or even without auxiliary data, remains an open research question. More importantly, in the era of big data, either individual agents or federated learning systems keep generating various versions of predictive models for a stable group of users~\cite{konevcny2016federated}, so if an adversary can collect multiple correlated models, they can potentially extract more comprehensive information of the original training data from them. However, to the best of our knowledge, no existing literature investigates the potential benefits of targeting multiple models simultaneously. 

We refer to a group of such correlated models as an \textit{ensemble}. Note we don't target situations in which models are \textit{trained to be used} as an ensemble for inference, but primarily other situations in which multiple models exist with overlapping predictions, such as common classes for classification models, which can be \textit{attacked} as an ensemble. Examples include different releases of a software product that comprise increasingly improved versions of a model, or different products that each have a different model for the same common purpose \cite{huang2017snapshot}. Another example includes Federated Learning, in which every participant continuously receives incrementally updated models including contributions from the training samples (e.g. images) of other participants. Our contributions are as follows:

\begin{enumerate}
\item We propose a model inversion attack method based on exploiting an ensemble of ML models, which demonstrates noticeable improvement of reconstruction performance over attacking a single model.
\item We investigate the impact of model diversity on the performance of ensemble inversion and propose the farthest model sampling (FMS) method to maximize the diversity of models in a collected ensemble.
\item We investigate common scenarios in which an adversary can potentially collect correlated models to construct an ensemble for inversion and determine class correspondence among multiple models.
\item We propose to utilize the richer information contained in the model output vector to provide better constraints for distinguishable attributes of the target identities.
\item We investigate the tradeoff between reconstructing realistic-looking images and distinguishable features, and present successful results for data-free ensemble inversion without any auxiliary data.
\end{enumerate}

\section{Related work}
Revealing sensitive attributes of training data is one of the major focuses of privacy attacks on ML models, which include membership attacks and model inversion attacks. The objective of membership attacks is to determine if one particular data sample has been used for training the model ~\cite{MembershipAttack}, while model inversion targets reconstruction of data following the same distribution as the original training data. To defend against various privacy attack methods, there exists a large body of work that formalizes the privacy notion and develops various techniques to prevent attacks from being successful. Differential Privacy (DP) is a widely discussed definition of privacy and various techniques have been developed to achieve it, which carefully randomize an algorithm making its outputs invariant to the presence or absence of any individual data sample~\cite{dwork2014algorithmic}. DP guarantees the protection of ML algorithms against attacks to reflect if a sample is from the training data of the trained model. However, DP does not explicitly defend against model inversion attacks, which focus on the recovery of attributes of the original training dataset.

Fredrikson \etal~\cite{fredrikson2014privacy} proposed the model inversion attack against ML algorithms, and proved that publishing predicted dosage amounts can cause leakage of personal genetic information in generalized linear regression. Then, Fredrikson \etal~\cite{fredrikson2015model} presented a model inversion attack on a face recognition model to reconstruct face images from the original training data. However, the recovered face images are blurry and the quality of reconstruction is limited by the complexity of the model. 

Kusano and Sakuma~\cite{kusano2018classifier} utilized a generative adversarial network (GAN) together with an auxiliary dataset to reconstruct images from classifiers. However, the quality of their reconstruction depends on an overlapping portion between original training data and the auxiliary dataset, which limited its practical use. Zhang \etal~\cite{zhang2020secret} also use a GAN for image reconstruction, but they have a blurred or partially blocked version of images of the original training identities as auxiliary dataset. Therefore, they can generate photorealistic images by applying image reconstruction constraints in a way similar to image inpainting. However, the assumption that a blurred version of original training data is available to the attacker is too strong in many scenarios. 

Our work differs from the existing literature because we identify that attackers can potentially obtain different versions of ML models trained on a common set of entities, and exploit model ensembles instead of one single model for a model inversion attack. We also quantitatively investigate the impact of model diversity in the ensemble on inversion performance and propose the farthest model sampling (FMS) technique to select a given number of candidate models for the construction of an ensemble for inversion. We believe that existing model inversion approaches can also benefit from integrating the techniques proposed in this paper. 

\section{Proposed method}
In this section, we present a novel model inversion framework, which extracts distinguishable attributes of training data from an ensemble of machine learning models. The overall diagram of the proposed method is illustrated in Fig.~\ref{fig:overview}. We will discuss GAN-based image reconstruction before introducing the ensemble inversion attack.

\subsection{GAN for training sample reconstruction}
Generative adversarial networks (GANs)~\cite{goodfellow2014generative} are successful generative methods, which can generate samples that follow the same distribution as the data they are trained on. GANs are usually composed of a generator $G$ and a discriminator $D$. Taking image generation as an example, $G$ receives a random noise vector $z$ as input and generates fake images $X_{fake}=G(z)$ trying to mimic the target distribution. Meanwhile $D$ is trained to distinguish between the real images $X_{real}$ (of the target distribution) and the fake images from the generator. During training, the generator $G$ is continuously updated according to the training error produced by $D$. In other words, $G$ is guided by $D$ to generate synthesized data $G(z)$, which follows the same distribution as the real images. On top of this vanilla GAN concept, class information can be added to the discriminator to guide $G$ to generate samples of particular classes~\cite{mirza2014conditional, odena2017conditional}. Typically, if every sample has a corresponding class label, $c \sim p_c$, $G$ can be trained to generate synthesized samples based on both $c$ and $z$, namely, $X_{fake} = G(c, z)$. Meanwhile, $D$ is used to constrain the probability distribution over the source (real or fake data) and an additional probability distribution over the class labels, $P(S | X), P(C | X) = D(X)$, where $S$ and $C$ represent source and class, respectively. There are two training objectives: the log-likelihood of the correct source $\mathcal{L}_S$ and the log-likelihood of the correct class $\mathcal{L}_C$~\cite{odena2017conditional}.
\begin{equation}   \label{eq1}
\begin{small}
\begin{aligned}
\mathcal{L}_{S}=\mathbb{E}[log P(S=r | X_{real})]+\mathbb{E}[log P(S=f | X_{fake})] \\
\mathcal{L}_{C}=\mathbb{E}[log P(C=c| X_{real})]+\mathbb{E}[log P(C=c| X_{fake})]\
\end{aligned}
\end{small}
\end{equation}

According to \cite{odena2017conditional}, $D$ is trained to maximize ${\mathcal{L}_S + \mathcal{L}_C}$ while $G$ is trained to maximize ${\mathcal{L}_C - \mathcal{L}_S}$. In our model inversion attack framework, the model under attack (MUA) is used to provide $P(C=c | X)$, which was achieved from learning from the original training data. Therefore, $G$ will be guided by the MUA to generate images with distinguishable attributes of the original training label $c$. Note that the MUA is frozen during the model inversion experiments.

\subsection{Ensemble inversion attack}
Mainstream DCNNs can be trained on arbitrarily large training data sets using stochastic training algorithms, such as SGD using mini batches. This makes DCNN models sensitive to initial random weights and statistical noise in the training dataset. This stochastic nature of learning algorithms introduces different versions of models, which pay attention to different features even though they are trained on the same dataset. Therefore, researchers usually use ensemble learning to reduce the variance, which is a straightforward way to improve results of discriminatively trained DCNNs~\cite{krizhevsky2012imagenet, wang2015unsupervised, zeiler2014visualizing}. 

This work is inspired by ensemble learning, but the concept of ensemble is different. For model inversion, attackers \textit{cannot} assume the models under attack are always trained with ensemble learning. However, they can potentially collect correlated models to construct an ensemble to attack. In other words, in the context of the ensemble inversion attack, the ensemble refers to a set of correlated models which attackers can collect from various sources, without the requirements that such collected models were \textit{trained} with ensemble learning. For example, researchers or companies will keep receiving new training data and train and release updates to existing models, which might be collected by an attacker and used as an ensemble. 

\begin{figure}
\centering
\includegraphics[width=0.8\linewidth]{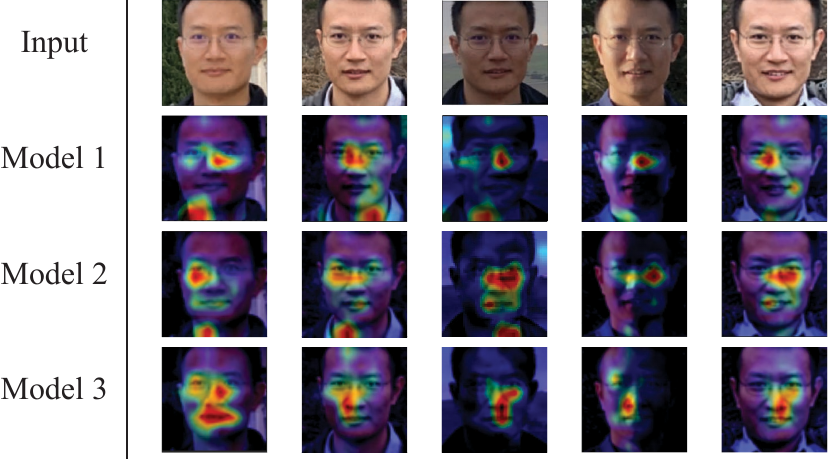}
\caption{Visualization with heatmaps generated by different models trained on the same dataset for a same set of inputs. Different models focus on different regions of a face image.}
\label{fig:heatmap}
\end{figure}

To the best of our knowledge, this property has never been explored in the field of model inversion, which naturally requires more information from models to improve the data reconstruction quality. The variance among ML models is visualized in Fig.~\ref{fig:heatmap}, in which we show heatmaps of one same set of images with five different face classifiers. It can be seen that the face classifier models pay attention to different parts of the face even though they were trained with the same data and loss. Therefore, more comprehensive features can be potentially extracted if they are combined together for a model inversion task.

For image classification tasks, the DCNN adopts the cross entropy loss in the training stage, which enforces the outputs to be close to ground truth labels of inputs. If the generator $G$ can produce synthetic (fake) samples $\{x_1 , x_2 , ... , x_n \}$, and assume they are passed to the $i$th MUA as inputs, we can obtain the outputs $\{y_1^i , y_2^i , ... , y_n^i\}$. Then, it is easy to calculate the predicted labels $\{t_1^i, t_2^i, ... , t_n^i\}$ as the one-hot representation of the $i$th MUA's prediction. Thus, inspired by \cite{chen2019data} we introduce the one-hot loss formulated as
\begin{equation} \label{eq4}
\mathcal{L}_{OH} =  \frac{1}{n}\sum_i{ \sum_j{H_{cross}(y_j^i, t_j^i)} },
\end{equation}
where $H_{cross}$ represents the cross entropy loss function. The purpose is to encourage $G$ to generate samples that will trigger sharp outputs of the $i$th MUA, instead of samples that can be potentially interpreted as multiple identities. Since the neuron activation in the output layer before softmax cross entropy also reflects the confidence of an MUA on its prediction, with $a^i_j$ representing the output activation of the $i$th MUA on the $j$th input sample, we propose a maximum response loss to guide $G$ to maximize the maximal activation of output neurons, where $k$ represents the index of elements in the activation vector $a^i_j$.

\begin{equation} \label{eq5}
\mathcal{L}_{MR} = -\frac{1}{n} \sum_i{ \sum_j{\max_{k}(a_j^i)} }
\end{equation}
 
The aforementioned losses are combined as Eq.~\ref{eq6} to constrain the generator $G$ to synthesize data which satisfies the predictions of all the MUAs in the ensemble.
\begin{equation} \label{eq6}
\resizebox{.9\hsize}{!}{$\mathcal{L}_{G} =  (\alpha_1 \mathcal{L}_{OH} + \alpha_2 \mathcal{L}_{MR} + \beta_1 \mathcal{L}_{class})/m + \beta_2 \mathcal{L}_{adv}$,}
\end{equation}
where $m$ is the number of models in the ensemble. Meanwhile, $\mathcal{L}_{adv}$ and $\mathcal{L}_{class}$ are the standard ACGAN losses ~\cite{odena2017conditional}, which correspond to $\mathcal{L}_S$ and $\mathcal{L}_C$ in Eq.~\ref{eq1}. $\mathcal{L}_{adv}$ is used to encourage $G$ to generate realistic-looking samples. $\mathcal{L}_{class}$ is applied to the outputs of the MUAs in our method instead of the output of the discriminator. We use a standard binary cross-entropy as discriminator loss $\mathcal{L}_{D}$. For all data-free experiments $\beta_2$ is set to 0. 

\subsection{Farthest model sampling}
\label{sec:fms}
In certain situations an attacker might have a large number of models to choose from, e.g. different snapshots or releases of an evolving model, and attacking an ensemble comprising all of these models might be computational prohibitive. Given the hypothesis that more dissimilar models will lead to more diverse perspectives on a given problem and therefore make an inversion attack more successful when included in an ensemble, we propose Farthest Model Sampling (FMS), a method to identify a diverse subset of models. Our method represents each model by a high-dimensional vector obtained from concatenated prediction vectors generated from a fixed set of randomized input data. We then use the farthest point sampling (FPS) method~\cite{qi2017pointnet++} widely used in applications such as point cloud sampling to (greedily) choose a subset of points of a desired size that have the farthest distance from each other. We use L2 distance between the aforementioned high-dimensional points to compute similarity of models. Compared to random sampling, it has better coverage of the entire point set given the same number of samples. 

\begin{figure}
\centering
\includegraphics[width=1.0\linewidth]{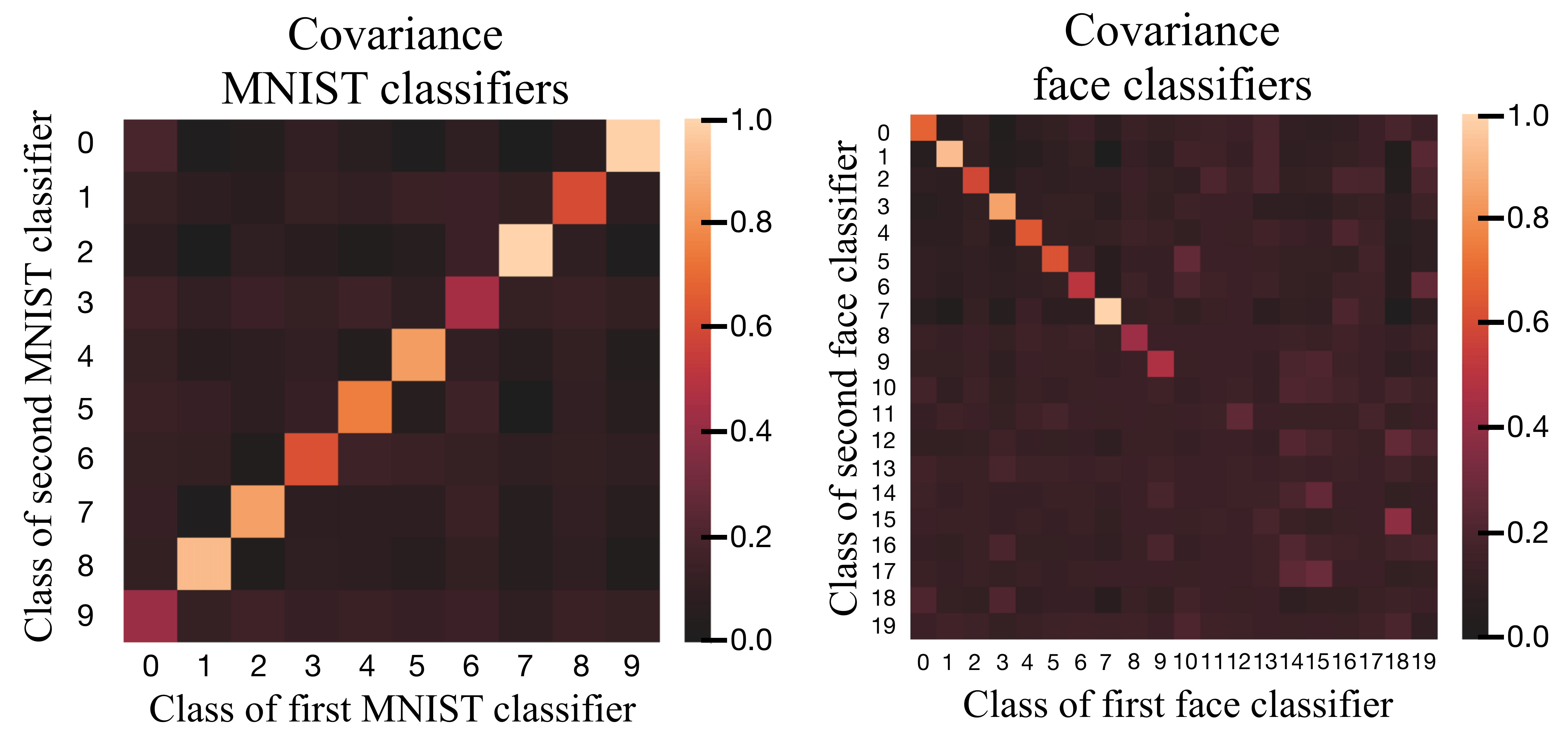} 
\caption{Left: The digits $0-9$ correspond to classes `0', `1' ... `9' for MNIST classifier 1, but classes `9', `8' ... `0' for MNIST classifier 2. Right: The face classifier 1 and 2 were trained with 20 identities, with only the first 10 identities shared.}
\label{fig:covariancematrix}
\end{figure}

\subsection{Class correspondence}
As mentioned earlier, one scenario of ensemble inversion is that the adversary (i.e. a participant of collaborative learning system) can collect different releases of a software product as correlated models to construct an ensemble under attack. In this case, the classes in the output vectors are likely aligned correctly among multiple classifiers. However, it remains a question how to find the correspondence between any two classifiers collected from different sources.  In other words, how can the adversary determine which class of one model corresponds to which class of another model? In some cases, the adversary can potentially utilize social network analysis methods \cite{knoke2019social} on other user data (i.e. ip address, membership number, age, gender, and browsing history) to estimate the correlation of classes. If only machine learning models are available, it is still possible to estimate the class correspondence by analyzing the responses of models with the same input stimulation. According to Fig.~\ref{fig:covariancematrix} (left), we pass EMNIST letters into two digit classifiers trained with MNIST but with different sequencing of classes in the prediction vectors. By analyzing the covariances of prediction vectors, we can identify the correct class correspondence between these two classifiers. Similarly for Fig.~\ref{fig:covariancematrix} (right), we can also correctly identify that the first ten classes are shared between these two face classifiers, based on the covariance between prediction vectors of two face classifiers on face images not used for training these classifiers.

\section{Experiments and results}
In this section we report on experimental results using different model and ensemble inversion attacks on digit and face classification datasets. We depend on attack accuracy and reconstruction visualization to evaluate the quality of the reconstructed images. To obtain attack accuracy, we train an evaluation classifier with an architecture different from the MUAs, on datasets involving the target identities for model inversion. This evaluation classifier should be independent of the ensemble inversion process in Fig.~\ref{fig:overview}, and it is used to predict the classes of the generated images after the ensemble inversion task is finished. A high attack accuracy indicates more leakage of private information about the target label. 

\subsection{Experiments on MNIST}
Our first experiments were implemented for the MNIST dataset~\cite{deng2012mnist}, which is composed of 28$\times$28 pixel images of handwritten digits from ten classes (0 to 9). The MNIST dataset contains 60,000 training images and 10,000 testing images. In our experiments, we first split the training dataset into four sub datasets. Because the data was shuffled before the partitioning, we assume the classes in each sub dataset are balanced. Then, we train a digit classifier for each subset, so that we can build three different ensembles with one, two and four MNIST classifiers, respectively. All the classifiers mentioned above were trained with Lenet-5 architecture~\cite{lecun2015lenet}, and they are used as the MUAs in our attacking setup.

For the ensemble inversion attack of MUAs, we investigate two common scenarios:
\begin{itemize}
 \item Data-free: We assume attackers don't have prior knowledge about the original training data and they cannot access any auxiliary datasets, so there is no need for the discriminator in Fig. \ref{fig:overview}.
 \item Auxiliary dataset: We assume attackers know the type of the task (i.e. digit classification), so they can utilize a similar dataset as auxiliary. To be fair, we chose the EMNIST letter dataset~\cite{cohen2017emnist} without any overlap with MNIST digits as the auxiliary. The motivation is to help $G$ to generate more realistic images (i.e. a character) more easily, while extracting distinguishable features for MNIST digits from the MUAs.
\end{itemize}

In order to evaluate the quality of the generated fake samples, we trained an independent evaluation classifier with a different backbone architecture, concretely ResNet-18~\cite{he2016deep}, based on the 10,000 testing images. After $G$ in Fig.~\ref{fig:overview} is trained to reconstruct the original training data of all MUAs, we make $G$ generate 10,000 images balanced among the ten classes and evaluate the attack accuracy of the independent evaluation digit classifier on the synthesized images.

\begin{table}[!b]
  \footnotesize
  \centering
  \caption{Attack accuracies of the evaluation digit classifier on the synthesized images. Table (a) and Table (b) are results for data-free and auxiliary data based experiments, respectively. }
  \subtable[Data-free experiments (attack accuracy)]{
     \begin{tabular}{|c|c|c|c|c|}
        \hline
        \multirow{2}{*}{}                 & \multirow{2}{*}{Loss type} & \multicolumn{3}{c|}{Num of models ($m$)} \\ \cline{3-5} 
                                  &                            & 1          & 2         & 4         \\ \hline\hline
        \multirow{3}{*}{Raw samples}      & Base                          & \cellcolor[rgb]{0.787,0.213,0} 0.213        & \cellcolor[rgb]{0.479,0.521,0}  0.521       & \cellcolor[rgb]{0.199,0.801,0}  0.801       \\ \cline{2-5} 
                                  			& $\mathcal{L}_{OH}$                         & \cellcolor[rgb]{0.778,0.222,0} 0.222        & \cellcolor[rgb]{0.293,0.707,0} 0.707       & \cellcolor[rgb]{0.191,0.809,0} 0.809       \\ \cline{2-5} 
                                  			& $\mathcal{L}_{OH}$ + $\mathcal{L}_{MR}$                   & \cellcolor[rgb]{0.38,0.620,0} 0.620        & \cellcolor[rgb]{0.101,0.899,0} 0.899       & \cellcolor[rgb]{0.097,0.903,0} 0.903       \\ \hline
        \multirow{3}{*}{Filtered samples} & Base                          & \cellcolor[rgb]{0.783,0.217,0} 0.217        & \cellcolor[rgb]{0.333,0.667,0} 0.667       & \cellcolor[rgb]{0.087,0.913,0} 0.913       \\ \cline{2-5} 
                                  & $\mathcal{L}_{OH}$                         & \cellcolor[rgb]{0.754,0.246,0} 0.246        & \cellcolor[rgb]{0.175,0.825,0} 0.825       & \cellcolor[rgb]{0.088,0.912,0} 0.912       \\ \cline{2-5} 
                                  & $\mathcal{L}_{OH}$ + $\mathcal{L}_{MR}$                   & \cellcolor[rgb]{0.343,0.657,0} 0.657        & \cellcolor[rgb]{0.1,0.900,0} 0.900       & \cellcolor[rgb]{0.078,0.922,0} 0.922       \\ \hline
        \end{tabular}
       }
  \subtable[Auxiliary dataset based experiments (attack accuracy)]{
     \begin{tabular}{|c|c|c|c|c|}
       \hline
       \multirow{2}{*}{}                 & \multirow{2}{*}{Loss type} & \multicolumn{3}{c|}{Num of models ($m$)} \\ \cline{3-5} 
                                  &                            & 1          & 2         & 4         \\ \hline\hline
       \multirow{3}{*}{Raw samples}      & Base                          & \cellcolor[rgb]{0.2,0.8,0} 0.800        &  \cellcolor[rgb]{0.198,0.802,0} 0.802       &  \cellcolor[rgb]{0.169,0.831,0} 0.831       \\ \cline{2-5} 
                                  & $\mathcal{L}_{OH}$                         & \cellcolor[rgb]{0.211,0.789,0} 0.789        & \cellcolor[rgb]{0.135,0.865,0} 0.865       & \cellcolor[rgb]{0.115,0.885,0} 0.885       \\ \cline{2-5} 
                                  & $\mathcal{L}_{OH}$ + $\mathcal{L}_{MR}$                   &  \cellcolor[rgb]{0.151,0.849,0} 0.849        &  \cellcolor[rgb]{0.104,0.896,0} 0.896       &  \cellcolor[rgb]{0.105,0.895,0} 0.895       \\ \hline
       \multirow{3}{*}{Filtered samples} & Base                          &  \cellcolor[rgb]{0.058,0.942,0} 0.942        &  \cellcolor[rgb]{0.049,0.951,0} 0.951       &  \cellcolor[rgb]{0.045,0.955,0} 0.955       \\ \cline{2-5} 
                                  & $\mathcal{L}_{OH}$                         &  \cellcolor[rgb]{0.054,0.946,0} 0.946        &  \cellcolor[rgb]{0.032,0.968,0} 0.968       &  \cellcolor[rgb]{0.041,0.959,0} 0.959       \\ \cline{2-5} 
                                  & $\mathcal{L}_{OH}$ + $\mathcal{L}_{MR}$                   &  \cellcolor[rgb]{0.143,0.857,0} 0.857        &  \cellcolor[rgb]{0.049,0.951,0} 0.951       &  \cellcolor[rgb]{0.021,0.979 ,0} 0.979       \\ \hline
       \end{tabular}
   }
   \label{table:mnist}
\end{table}

\begin{figure*}[!t]
\centering
\includegraphics[width=1.0\linewidth]{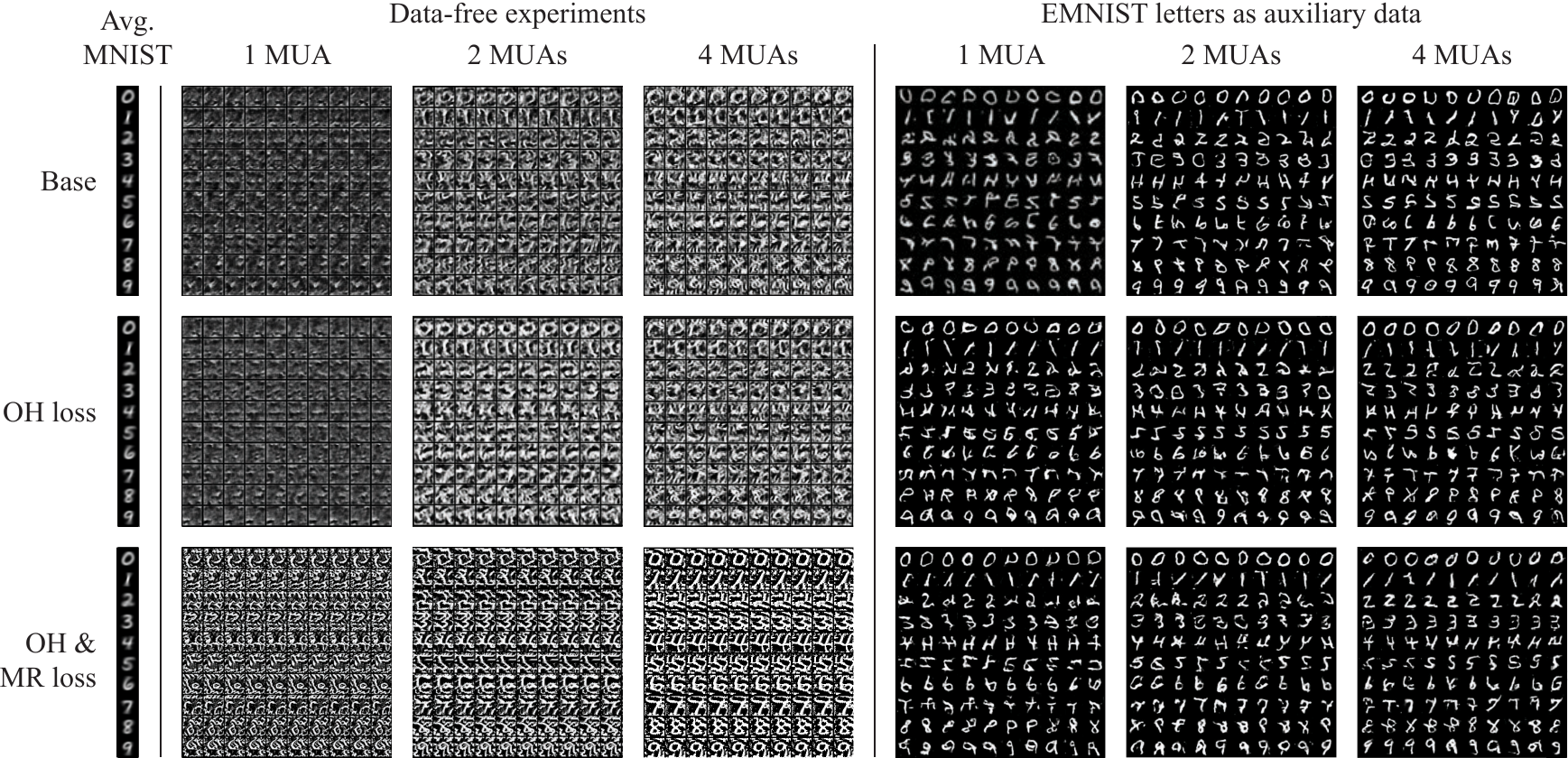}
\caption{Comparison of reconstructed samples from the data-free and auxiliary data based setups. Both increasing the number of models under attack (MUAs) and applying the proposed losses can improve the quality of reconstruction. The auxiliary data-based experiments can lead to more realistic samples, but data-free experiments also generates visually recognizable patterns using the proposed method.}
\label{fig:mnist_datafree_and_aux}
\end{figure*}

The attack accuracies are summarized in Table~\ref{table:mnist}. Table~\ref{table:mnist}(a) shows the results of the data-free experiments without any auxiliary data, while Table~\ref{table:mnist}(b) shows the experiments with EMNIST letters as auxiliary data. In each subtable, we evaluate two types of synthesized data, one is 10k raw images directly obtained from $G$, while the other is 10k filtered samples. For the filtered samples, we first generated 100k raw samples from $G$, but only kept the 10\% of them with highest maximal output activations for all MUAs. This filtering process is entirely automatic based on prediction results of MUAs, without any human interference. The filtered samples are shown in Fig.~\ref{fig:mnist_datafree_and_aux}.

According to Table~\ref{table:mnist}, the experiments with the EMNIST letter auxiliary dataset (Table~\ref{table:mnist}(b)) demonstrate higher accuracies than the data-free setup (Table~\ref{table:mnist}(a)). Meanwhile, better accuracies can be obtained when the number of MUAs in the ensemble is increased from 1 to 4 for most experiments, which confirms our earlier hypothesis that the ensemble attack tends to extract better features than a single model attack. We repeated these experiments for different loss types listed in the second column, where ``Base'' means the experiments did not involve $\mathcal{L}_{OH}$ or $\mathcal{L}_{MR}$, which can already achieve decent attack accuracies with ensembles. But it is obvious that $\mathcal{L}_{OH}$ and $\mathcal{L}_{MR}$ can further boost the performance in most cases. The corresponding weights are set to $\alpha_1$=200 and $\alpha_2$=0.0001. Finally, applying maximal activation-based filtering on the samples also proves to be an effective way to collect good samples. Although there exist some exceptions, we can conclude that overall above-mentioned techniques improve the attacking performance. 

\begin{figure}[t]
\centering
\includegraphics[width=1.0\linewidth]{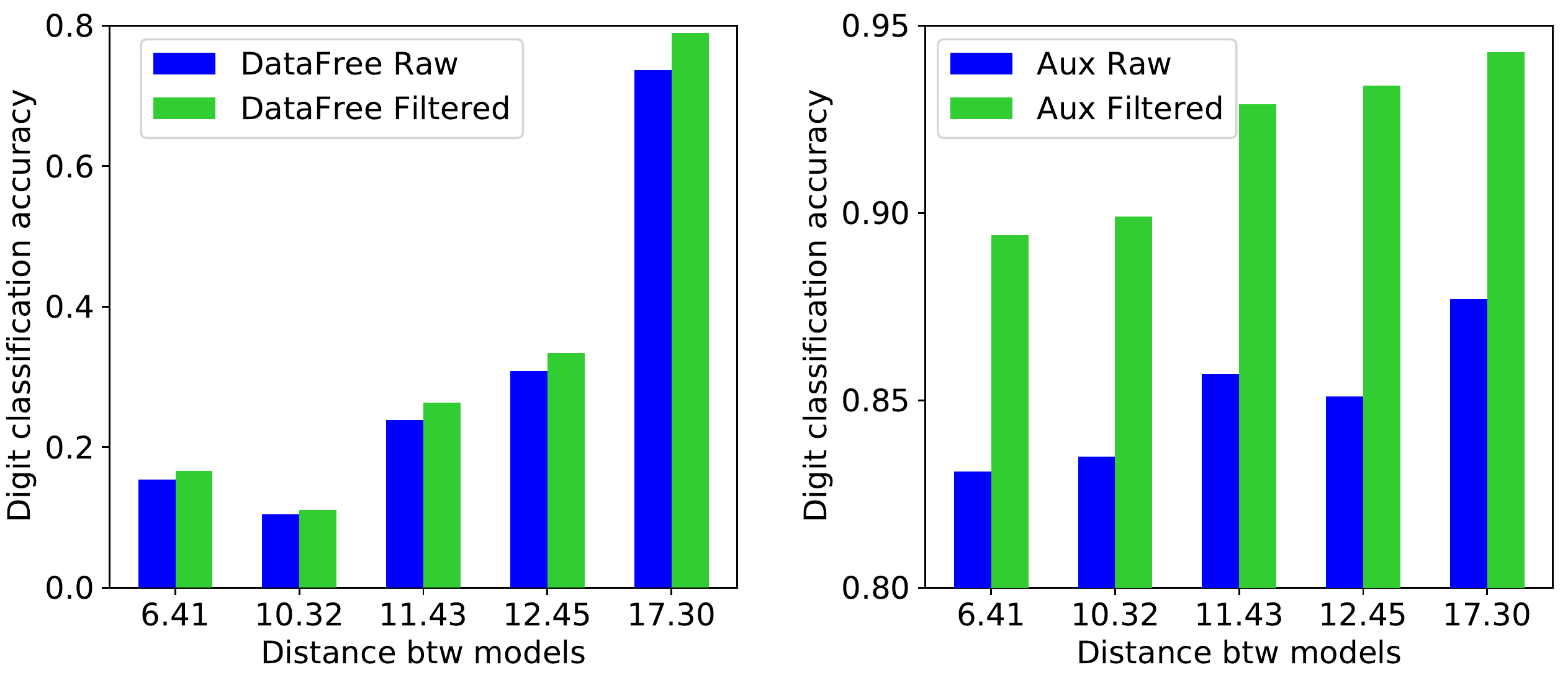}
\caption{Attack accuracy on images from generators trained with different two-model ensembles with increasingly higher distance between the two MUAs. The bar chart on the left shows the data-free experiments, the left shows results with auxiliary dataset.}
\label{fig:mnist_distance_bars}
\end{figure}

On top of different initialization of models, we believe the diversity of models within an ensemble plays an important role here. Intuitively, if the MUAs in the ensemble would all share exactly the same parameters, the ensemble inversion might not be more effective than single model inversion. Fig.~\ref{fig:mnist_distance_bars} illustrates how diversity of MUAs affects the ensemble inversion performance. In this experiment, we trained 200 digit classifiers (Lenet-5), and the training of each classifier is based on 30,000 randomly sampled images from the MNIST training set. To obtain such a large number of trained models, unlike the data partitioning in the previous experiment, we allow data overlap among these 200 classifiers. We first made them predict for the same set of 10,000 noise samples, so each digit classifier provides a 10,000 $\times$ 10 prediction vector. Then, we calculated the L2 distance between prediction vectors of any two of these digit classifiers, which was finally sorted based on the L2 distance. For Fig.~\ref{fig:mnist_distance_bars}, we sampled five pairs of MUAs whose L2 distances range from 6.409 to 17.302 to evaluate ensemble inversion performance. We can clearly see the positive correlation between the L2 distance between the models and the attack accuracy.

\begin{figure}[thb]
\centering
\includegraphics[width=1.0\linewidth]{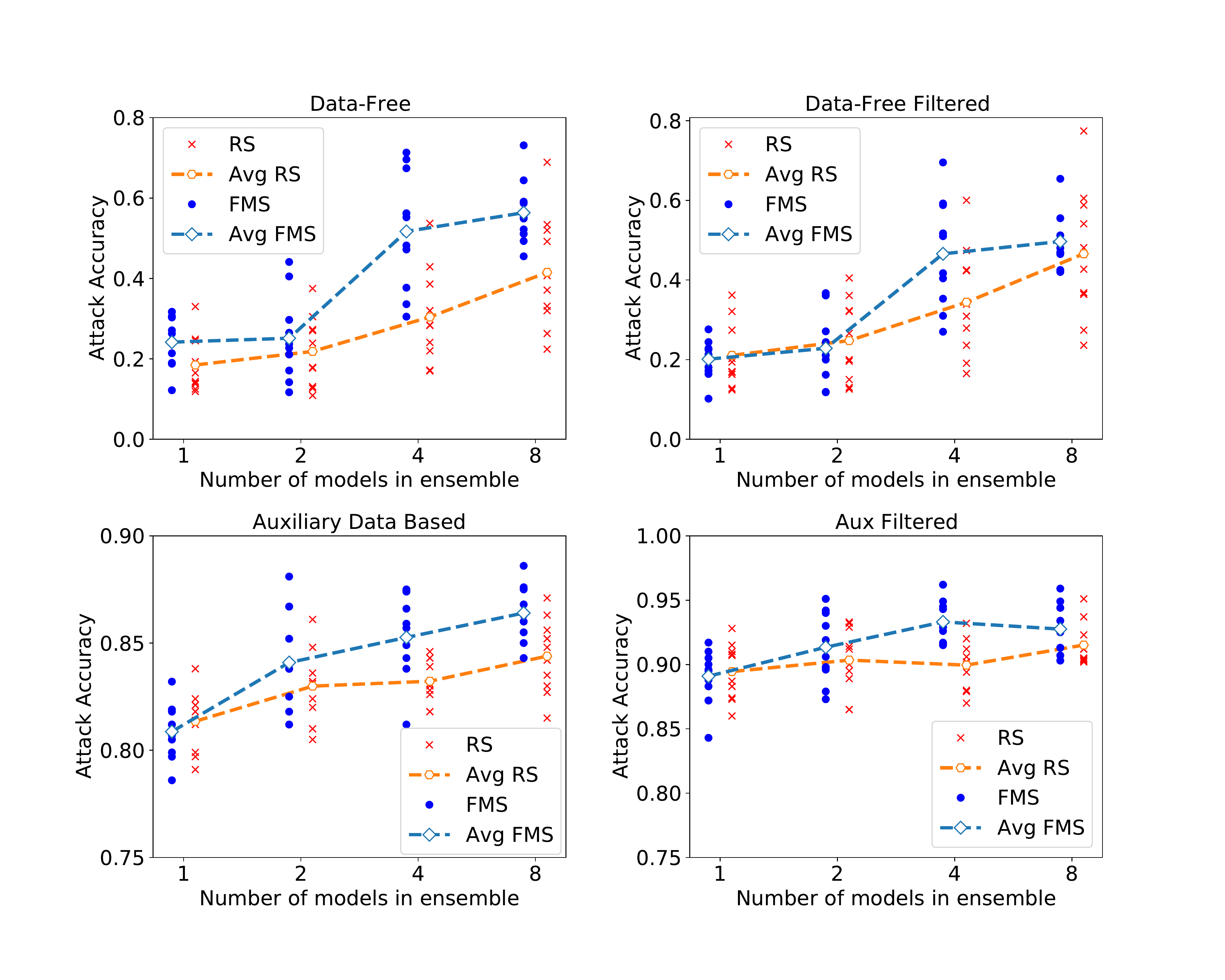}
\caption{The impact of using different numbers of models on the ensemble inversion performance. Two sampling methods are used to select models for the ensembles: the proposed FMS method and simple random sampling (RS).}
\label{fig:fms_result}
\end{figure}

To evaluate how the selection of models to form an ensemble impacts the reconstruction performance, we did the inversion experiments for ensembles of size 1, 2, 4 and 8 sampled from 200 models with two different methods: our proposed FMS method, explained in section~\ref{sec:fms}, and random sampling (RS). The results are illustrated in Fig.~\ref{fig:fms_result}. The blue dots and curves in Fig.~\ref{fig:fms_result} correspond to the proposed FMS method, while the red dots and orange curves are for the RS based experiments. For each ensemble size and model sampling method, we performed ten experiments with different weight initializations. By looking at the averaged attack accuracies (Avg. RS and Avg. FMS), we can conclude that using more target models for ensemble inversion in most cases improves the attack accuracy on the reconstructed data for both model sampling methods. Furthermore, we observe that FMS tends to provide better performance than RS if the ensemble size is larger than one. This means that explicit constraints on the model diversity can help to achieve more successful ensemble inversion.

Fig.~\ref{fig:fms_result} already covers various diversities of models under attack, but it is worth to specifically investigate a common scenario that an attacker samples snapshots of one model during training and constructs an ensemble for inversion attack. This scenario is very realistic and it does not intuitively lead to the best diversity.  However, according to Table~\ref{table:rebuttal_distance_metric}, the improvement of attack accuracy due to ensemble inversion still exists. What is more, we also used distance metrics similar to~\cite{zhang2020secret} to evaluate how similar the reconstructed images are to the original training data.  We extract features from the penultimate layer of the evaluation classifier (EVA) and define \textit{feature distance} as the L2 distance between features of reconstructed images and the centroid of the features of training images of the same class. Similarly, \textit{kNN distance} is defined as the L2 distance from the reconstructed images to the closest training image of the same class in the feature space. We observe that attacking larger ensembles decreases both feature and kNN distances between reconstructed images and original training images. \begin{table}[thb]
\footnotesize 
\centering
\caption{Results using updated versions of an MNIST classifier sampled from multiple epochs as ensemble.}
\begin{tabular}{|c|c|c|c|c|c|}
\hline
                                                      & \begin{tabular}[c]{@{}c@{}}Attack\\ Acc$\uparrow$\end{tabular} & \begin{tabular}[c]{@{}c@{}}Feat Dist \\ (EVA)$\downarrow$\end{tabular} & \begin{tabular}[c]{@{}c@{}}kNN Dist \\ (EVA)$\downarrow$\end{tabular} & \begin{tabular}[c]{@{}c@{}}Feat Dist \\ (CIF)$\downarrow$ \end{tabular} & \begin{tabular}[c]{@{}c@{}}kNN Dist \\ (CIF)$\downarrow$\end{tabular} \\ \hline
\begin{tabular}[c]{@{}c@{}}1 MUA\end{tabular}  & 0.898                                                & 2,760.5                                  & 1,221.3                                & 6,291.8                                    & 3,418.4                                    \\ \hline
\begin{tabular}[c]{@{}c@{}}2 MUAs\end{tabular} & 0.932                                               & 2,594.9                                  & 1,128.3                                & 6,014.1                                    & 3,316.9                                    \\ \hline
\begin{tabular}[c]{@{}c@{}}4 MUAs\end{tabular} & 0.950                                                & 2,480.5                                 & 1,116.2                                & 6,004.7                                    & 3,092.5                                     \\ \hline
\end{tabular}
\label{table:rebuttal_distance_metric}
\end{table}

\begin{table}[thb]
\footnotesize 
\centering
\caption{Comparison of attack accuracy, feature and kNN distances when performing inversion on 1 and 5 face classifiers.}
\begin{tabular}{|c|c|c|c|c|c|}
\hline
                                                      & \begin{tabular}[c]{@{}c@{}}Attack\\ Acc$\uparrow$\end{tabular} & \begin{tabular}[c]{@{}c@{}}Feat Dist \\ (EVA)$\downarrow$\end{tabular} & \begin{tabular}[c]{@{}c@{}}kNN Dist \\ (EVA)$\downarrow$\end{tabular} & \begin{tabular}[c]{@{}c@{}}Feat Dist \\ (CIF)$\downarrow$\end{tabular} & \begin{tabular}[c]{@{}c@{}}kNN Dist \\ (CIF)$\downarrow$\end{tabular} \\ \hline
\begin{tabular}[c]{@{}c@{}}1 MUA\end{tabular}  & 0.683                                                & 467.4                                                      & 97.3                                                      & 4,477.2                                                     & 1,663.9                                                    \\ \hline
\begin{tabular}[c]{@{}c@{}}5 MUAs\end{tabular} & 0.894                                                & 329.9                                                      & 60.8                                                      & 4,167.1                                                     & 1,607.3                                                    \\ \hline
\end{tabular}
\label{table:Face_Attack}
\end{table}

To rule out the influence of using the same type of training data to obtain the evaluation classifier (EVA), we further computed those same metrics on the features of a CIFAR100 classifier (CIF) instead of MNIST (or a face classifier for later experiments), which gives a better indication of general image similarity as opposed to similarity to a digit (or face) classification network. The results in Table~\ref{table:rebuttal_distance_metric} again show the same correlation, indicating that larger ensembles lead to images that look more similar to the training images.

\subsection{Experiments on face classification models} 
For the ensemble inversion attack of face classifiers, we explore the scenario in Fig.~\ref{fig:covariancematrix} (right), in which attackers have access to multiple prediction models trained on different groups of identities and the shared identities among these groups are the target. It is a common scenario because attackers might usually be interested in the identities who are members of multiple organizations. 

In reality, the group size can vary and the architecture of corresponding classifier can be different. In this case, the ensemble inversion technique is expected to boost the inversion performance since different identity groups can encourage the classifiers to extract different distinguishable features of the shared identities. Note that we focus on the shared identities among face classifiers, while ignoring the other identities during ensemble inversion.

\begin{figure*}[!h]
\centering
\includegraphics[width=0.95\linewidth]{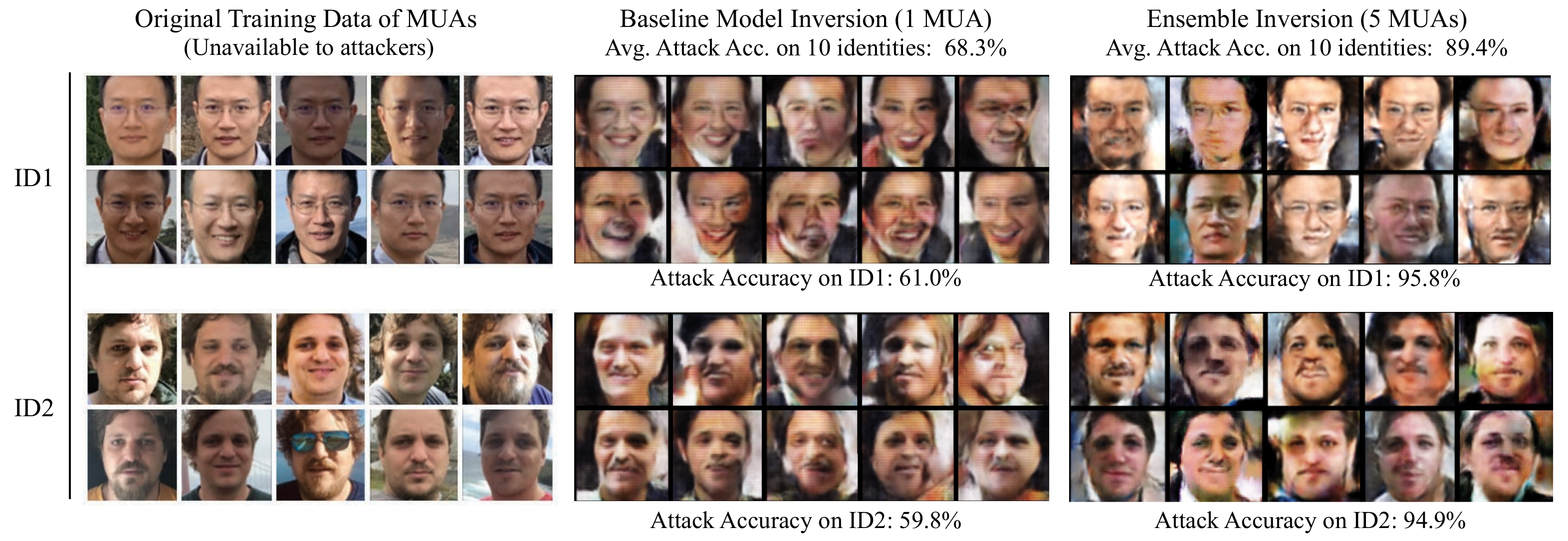}
\caption{For baseline model inversion and the proposed ensemble inversion setups, we generated 1,000 samples for each target ID. We visualize two target IDs (i.e. ID1 and ID2) and show 10 samples for each identity for each experiment. The accuracy values are generated by a separate evaluation classifier on the reconstructed images.}
\label{fig:Face_Attack_RGB}
\end{figure*}
\begin{figure}[thb]
\centering
\includegraphics[width=0.95\linewidth, height=0.5\linewidth]{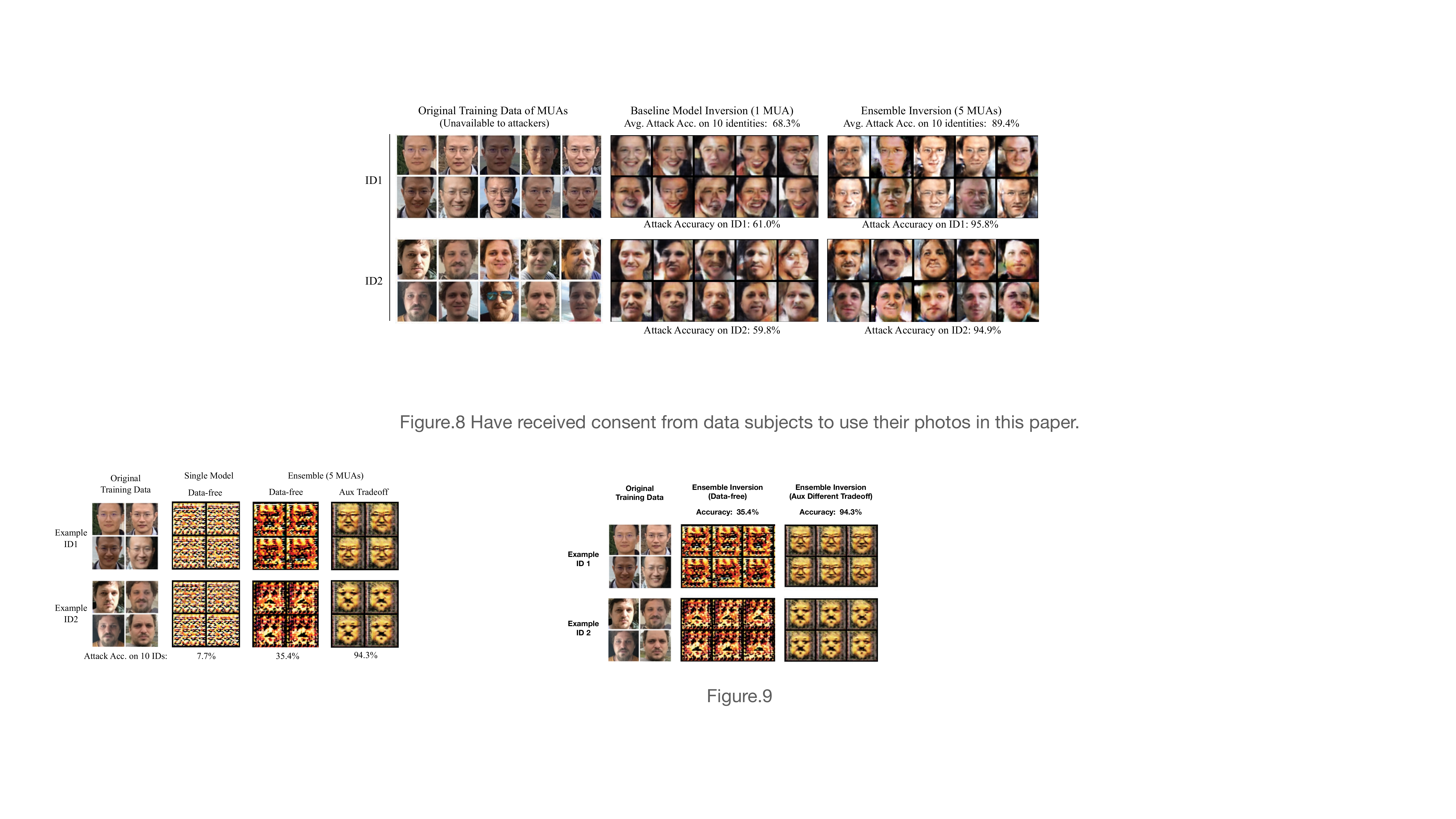}
\caption{Comparison of reconstruction results of two data-free experiment (single model vs. ensemble inversion) and auxiliary data based reconstruction with a new weight balancing ($\alpha_1$=500, $\alpha_2$=0.01, $\beta_1$=0.9, $\beta_2$=0.1).}
\label{fig:Face_Attack_Datafree}
\end{figure}

We trained face classifiers with ResNet-34~\cite{he2016deep} as backbone on the VGGFace2 dataset~\cite{cao2018vggface2}, which contains 3.31 million images of 9,131 identities. In order to prevent overlap between training identities and auxiliary identities, we split VGGFace2 into two parts: 500 identities for training MUAs, and the remaining identities for auxiliary data. 

In our experiment, we trained five face classifiers for 100 identities based on a ResNet-34 backbone. The first ten identities are the same for all five classifiers, and the remaining identities were randomly sampled from the 500 identities we set aside for training from VGGFace2. To evaluate the quality of the reconstructed images for the ten shared identities, we trained an independent evaluation face classifier with VGG16 backbone~\cite{simonyan2014very} on these ten target identities and another 60 identities, which are not used for training any of these ResNet-34 classifiers. In this section, all attack accuracies on the reconstructed images are obtained from the independent evaluation face classifier. While we report quantitative results for 10 identities, we show visual results only for two identities from whom we received consent to reproduce their photos.

Fig.~\ref{fig:Face_Attack_RGB} compares the baseline model inversion experiment for one single target model without the proposed losses (i.e. $\mathcal{L}_{OH}$ and $\mathcal{L}_{MR}$) and the ensemble inversion experiment on five target models with the proposed losses applied (the loss weights are set to $\alpha_1$=200, $\alpha_2$=0.005, $\beta_1=\beta_2=0.5$). It can be seen that the faces reconstructed by the ensemble inversion experiment show more distinguishable features of the target identities (cf. training data) and the corresponding attack accuracy of the evaluation classifier is higher than that of the baseline experiment. Both overall and per-class attack accuracies are improved, which is consistent with the qualitative visual results.  Note that our main focus is to investigate techniques to help extracting distinguishable features of target identities without any overlap between the target identities and the auxiliary dataset instead of high-quality face synthesis with GANs. The attack accuracies and distance metrics are summarized in Table~\ref{table:Face_Attack}.

According to Fig.~\ref{fig:Face_Attack_Datafree},  ensemble inversion introduces noticeable improvement over single model inversion for data-free experiments. Although the reconstructed images can be coarse and the corresponding accuracy is lower than that of auxiliary data-based experiments, some patterns with recognizable facial features are still extracted successfully. The face images on the right column of Fig.~\ref{fig:Face_Attack_Datafree} are from an auxiliary data-based experiment, but with a smaller weight on $L_{adv}$ and higher weights on the other losses. Although these results are not as photorealistic as the images in Fig.~\ref{fig:Face_Attack_RGB} and some mode collapse is observed, the generator tends to learn the most recognizable features (or silhouette) of the target faces, and the distinguishable features of the target identities have been captured successfully. Therefore, the overall accuracy is higher than that of the photorealistic reconstructions in Fig.~\ref{fig:Face_Attack_RGB}, and balancing among different types of constraints allows the model (or ensemble) inversion task to achieve different tradeoffs.

\section{Conclusions} 
In this work, we propose the ensemble inversion technique which leverages the diversity of an ensemble of ML models to boost model inversion performance. On top of that, we analyzed common scenarios of obtaining variance of target models and thoroughly investigate how the diversity of target models can influence the ensemble inversion performance. In addition, we introduce one-hot loss and maximum output activation loss, which lead to further improvement on the quality of generated samples. Meanwhile, filtering out generated samples with low maximum activations of the models under attack, can further make the reconstructions more recognizable. When applying the proposed techniques, the MNIST digit reconstruction accuracy is improved by 70.9\% for the data-free experiment and 17.9\% for the auxiliary data based experiment. The face reconstruction accuracy is improved by 21.1\% over the baseline experiment. The motivation of this work is to provide a systemic analysis of the potential impact of the proposed techniques on model inversion. For future work, we will focus on the development of the corresponding defense mechanisms against such ensemble inversion attacks.

{\small
\bibliographystyle{ieee_fullname}
\bibliography{egbib}

\begin{thebibliography}{10}\itemsep=-1pt

\bibitem{cao2018vggface2}
Qiong Cao, Li Shen, Weidi Xie, Omkar~M Parkhi, and Andrew Zisserman.
\newblock {VGGFace2}: A dataset for recognising faces across pose and age.
\newblock In {\em Proc. 13th IEEE International Conference on Automatic Face \&
  Gesture Recognition (FG 2018)}, pages 67--74. IEEE, 2018.

\bibitem{chen2019data}
Hanting Chen, Yunhe Wang, Chang Xu, Zhaohui Yang, Chuanjian Liu, Boxin Shi,
  Chunjing Xu, Chao Xu, and Qi Tian.
\newblock Data-free learning of student networks.
\newblock In {\em \ICCV}, pages 3514--3522, 2019.

\bibitem{cohen2017emnist}
Gregory Cohen, Saeed Afshar, Jonathan Tapson, and Andre Van~Schaik.
\newblock {EMNIST}: Extending {MNIST} to handwritten letters.
\newblock In {\em Proc. International Joint Conference on Neural Networks
  (IJCNN)}, pages 2921--2926. IEEE, 2017.

\bibitem{deng2012mnist}
Li Deng.
\newblock The {MINST} database of handwritten digit images for machine learning
  research [best of the web].
\newblock {\em IEEE Signal Processing Magazine}, 29(6):141--142, 2012.

\bibitem{dwork2014algorithmic}
Cynthia Dwork and Aaron Roth.
\newblock The algorithmic foundations of differential privacy.
\newblock {\em Foundations and Trends in Theoretical Computer Science},
  9(3-4):211--407, 2014.

\bibitem{fredrikson2015model}
Matt Fredrikson, Somesh Jha, and Thomas Ristenpart.
\newblock Model inversion attacks that exploit confidence information and basic
  countermeasures.
\newblock In {\em Proc. 22nd ACM SIGSAC Conference on Computer and
  Communications Security}, pages 1322--1333, 2015.

\bibitem{fredrikson2014privacy}
Matthew Fredrikson, Eric Lantz, Somesh Jha, Simon Lin, David Page, and Thomas
  Ristenpart.
\newblock Privacy in pharmacogenetics: An end-to-end case study of personalized
  warfarin dosing.
\newblock In {\em Proc. 23rd {USENIX} Security Symposium ({USENIX} Security
  14)}, pages 17--32, 2014.

\bibitem{goodfellow2014generative}
Ian Goodfellow, Jean Pouget-Abadie, Mehdi Mirza, Bing Xu, David Warde-Farley,
  Sherjil Ozair, Aaron Courville, and Yoshua Bengio.
\newblock Generative adversarial nets.
\newblock In {\em \NIPS}, pages 2672--2680, 2014.

\bibitem{he2016deep}
Kaiming He, Xiangyu Zhang, Shaoqing Ren, and Jian Sun.
\newblock Deep residual learning for image recognition.
\newblock In {\em \CVPR}, pages 770--778, 2016.

\bibitem{huang2017snapshot}
Gao Huang, Yixuan Li, Geoff Pleiss, Zhuang Liu, John~E Hopcroft, and Kilian~Q
  Weinberger.
\newblock Snapshot ensembles: Train 1, get m for free.
\newblock {\em arXiv preprint arXiv:1704.00109}, 2017.

\bibitem{knoke2019social}
David Knoke and Song Yang.
\newblock {\em Social network analysis}.
\newblock Sage Publications, 2019.

\bibitem{konevcny2016federated}
Jakub Kone{\v{c}}n{\`y}, H~Brendan McMahan, Felix~X Yu, Peter Richt{\'a}rik,
  Ananda~Theertha Suresh, and Dave Bacon.
\newblock Federated learning: Strategies for improving communication
  efficiency.
\newblock {\em arXiv preprint arXiv:1610.05492}, 2016.

\bibitem{krizhevsky2012imagenet}
Alex Krizhevsky, Ilya Sutskever, and Geoffrey~E Hinton.
\newblock Imagenet classification with deep convolutional neural networks.
\newblock In {\em \NIPS}, pages 1097--1105, 2012.

\bibitem{kusano2018classifier}
Kosuke Kusano and Jun Sakuma.
\newblock Classifier-to-generator attack: Estimation of training data
  distribution from classifier, 2018.

\bibitem{lecun2015lenet}
Yann LeCun.
\newblock {LeNet-5}, convolutional neural networks.
\newblock {\em http://yann.lecun.com/exdb/lenet}, 20(5):14, 2015.

\bibitem{mirza2014conditional}
Mehdi Mirza and Simon Osindero.
\newblock Conditional generative adversarial nets.
\newblock {\em arXiv preprint arXiv:1411.1784}, 2014.

\bibitem{odena2017conditional}
Augustus Odena, Christopher Olah, and Jonathon Shlens.
\newblock Conditional image synthesis with auxiliary classifier {GANs}.
\newblock In {\em \ICML}, pages 2642--2651, 2017.

\bibitem{qi2017pointnet++}
Charles~Ruizhongtai Qi, Li Yi, Hao Su, and Leonidas~J Guibas.
\newblock Pointnet++: Deep hierarchical feature learning on point sets in a
  metric space.
\newblock In {\em \NIPS}, pages 5099--5108, 2017.

\bibitem{MembershipAttack}
Shokri Reza, Stronati Marco, Song Congzheng, and Shmatikov Vitaly.
\newblock Membership inference attacks against machine learning models.
\newblock In {\em Proc. IEEE Symposium on Security and Privacy}, pages 3--18,
  2017.

\bibitem{simonyan2014very}
Karen Simonyan and Andrew Zisserman.
\newblock Very deep convolutional networks for large-scale image recognition.
\newblock {\em arXiv preprint arXiv:1409.1556}, 2014.

\bibitem{wang2015unsupervised}
Xiaolong Wang and Abhinav Gupta.
\newblock Unsupervised learning of visual representations using videos.
\newblock In {\em \ICCV}, pages 2794--2802, 2015.

\bibitem{zeiler2014visualizing}
Matthew~D Zeiler and Rob Fergus.
\newblock Visualizing and understanding convolutional networks.
\newblock In {\em \ECCV}, pages 818--833. Springer, 2014.

\bibitem{zhang2020secret}
Yuheng Zhang, Ruoxi Jia, Hengzhi Pei, Wenxiao Wang, Bo Li, and Dawn Song.
\newblock The secret revealer: generative model-inversion attacks against deep
  neural networks.
\newblock In {\em \CVPR}, pages 253--261, 2020.

\end{thebibliography}
}

\end{document}